\documentclass[10pt,twocolumn,letterpaper]{article}

\usepackage[pagenumbers]{cvpr}

\definecolor{cvprblue}{rgb}{0.21,0.49,0.74}
\usepackage[pagebackref,breaklinks,colorlinks,allcolors=cvprblue]{hyperref}

\usepackage{algorithm}
\usepackage{algorithmic}
\usepackage{newfloat}
\usepackage{listings}
\usepackage{multirow}
\usepackage{tabularx}

\def\confName{CVPR}
\def\confYear{2027}

\title{Preserving Pretrained CFG Structure in RL for Diffusion Models\\ with Conditional Contrast Regularization}

\author{%
Yunfei Zheng\textsuperscript{1,2,*} \quad
Xiaolong Wang\textsuperscript{2,*} \quad
Dandan Zheng\textsuperscript{1,2} \quad
Xianjun Deng\textsuperscript{1}\\[0.35em]
Shenghao Liu\textsuperscript{1} \quad
Wei Xiang\textsuperscript{3} \quad
Yuqian Li\textsuperscript{2} \quad
Laurence Tianruo Yang\textsuperscript{1,\textdagger}\\[0.7em]
\textsuperscript{1}Huazhong University of Science and Technology\\[0.15em]
\textsuperscript{2}Ant Group \quad
\textsuperscript{3}Central China Normal University\\[0.55em]
{\small \textsuperscript{*}Equal contribution \quad
\textsuperscript{\textdagger}Corresponding author.}%
}

\begin{document}
\maketitle

\begin{abstract}

Reinforcement Learning (RL) has become an important component in the post-training of diffusion models.
In practice, however, standard RL often improves the target reward without reliably preserving holistic generation quality, leading to unstable reward--quality trade-offs that we view as a form of reward hacking.
Since Classifier-Free Guidance (CFG) is known to improve visual clarity, prompt alignment, and perceptual quality in generation, we investigate whether its underlying structure can serve as a useful prior during RL.
We present a unified geometric view that interprets both CFG and RL as correction vectors in a common prediction space, and identify the projection of the RL-induced correction onto the pretrained CFG correction as a useful diagnostic of quality-related structural preservation.
Geometric analysis and rotation intervention further show that the pretrained CFG direction carries information useful for generation quality that reward-driven RL updates do not necessarily preserve.
Based on this insight, we propose Conditional Contrast Regularization (CCR), which uses the frozen reference model's conditional contrast as a local structural prior during RL.
Experiments across multiple reward settings and backbones show that CCR more consistently preserves generation quality while maintaining competitive reward, yielding a better reward--quality trade-off than standard RL and generic regularization methods.

\end{abstract}

\section{Introduction}

\begin{figure}[t]
    \centering
    \includegraphics[width=0.95\columnwidth]{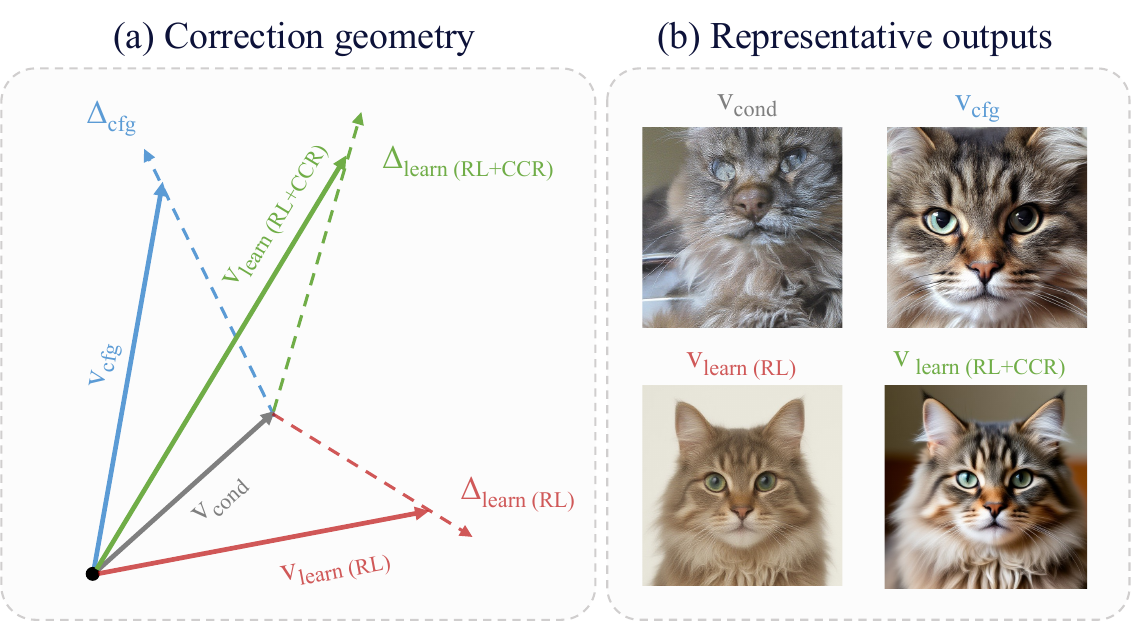}
    \caption{
    Conceptual illustration of CCR. (a) RL may produce a correction that deviates from the CFG correction direction, while CCR helps preserve the CFG-aligned component. (b) Representative images generated without CFG, with CFG, after RL fine-tuning, and after RL+CCR fine-tuning.
    }
    \label{fig:conceptual_overview}
\end{figure}

Reinforcement Learning (RL) has become an important post-training paradigm for diffusion models~\cite{sohl2015deep,ho2020denoising,song2020score} and flow-based models~\cite{esser2024scaling,wu2025qwen}, enabling optimization toward preference-based or task-specific objectives beyond likelihood training~\cite{black2023training,fan2023dpok,wallace2024diffusion,zheng2025diffusionnft}.
In practice, however, higher reward does not necessarily translate into better overall generation quality.
Standard RL can substantially improve the target reward, while visual clarity, prompt alignment, and perceptual quality often exhibit unstable or inconsistent behavior across reward settings and backbones~\cite{he2025gardo,mao2025image}.
We view this reward--quality mismatch as a form of reward hacking~\cite{gao2023scaling,dandapanthula2026we}, where RL increasingly favors reward-visible signals without reliably preserving the broader qualities desired in conditional image generation.

Existing approaches mainly address this issue through regularization toward a reference policy or data distribution~\cite{fan2023dpok,liu2025flow,ye2025data,he2025gardo}, or by improving the reward side through stronger reward models and reward design~\cite{liu2026beyond,zhao2026marble,zhang2026drm}.
These directions are important, but they do not directly answer a more structural question: during RL, which ingredients of the pretrained model should be preserved in order to maintain high-quality generation?
In particular, generic regularization can reduce policy drift in magnitude, yet it does not explicitly preserve the quality-related structure that may be lost or distorted during RL.
As a result, simply constraining deviation from a reference model is not sufficient to reliably maintain the reward--quality trade-off.

A natural clue comes from Classifier-Free Guidance (CFG)~\cite{ho2022classifier}, which remains one of the most effective mechanisms for improving conditional generation in diffusion models.
In pretrained models, CFG is widely observed to improve visual clarity, prompt alignment, and overall perceptual quality~\cite{bradley2024classifier}.
This suggests that the conditional contrast underlying CFG may encode useful structural information for high-quality generation.
Meanwhile, recent post-training results show that fine-tuning can partially replace explicit inference-time guidance~\cite{chen2024toward,chen2025visual,tang2025diffusion,zheng2025direct,li2026mclr}.
In particular, DiffusionNFT~\cite{zheng2025diffusionnft} suggests that, even in a CFG-free setting, RL alone can outperform a CFG-enabled pretrained model on reward metrics, indicating that RL may partially substitute for external guidance. 
However, higher reward does not necessarily mean that the quality-related benefits originally provided by CFG are retained.
This motivates us to ask whether CFG can provide a structural prior for RL, rather than serving only as an inference-time technique.

To study this question, we adopt a unified geometric view in which both CFG and RL are interpreted as \emph{corrections} to the pretrained conditional prediction, as illustrated in Fig.~\ref{fig:conceptual_overview}.
Under this view, CFG provides a conditional contrast correction, whereas RL provides a learned correction.
This places CFG and RL in a common prediction space and allows their relationship to be analyzed directly.
We further identify the projection of the RL-induced correction onto the pretrained CFG correction as a useful diagnostic of whether RL preserves a quality-related structural component.
Geometric analysis and an inference-time rotation intervention show that the pretrained CFG direction carries information beneficial to generation quality, while reward-driven RL updates do not necessarily retain this component.

These observations suggest that the key challenge is not merely to limit policy drift, but to preserve an appropriate component of the RL update along a quality-relevant direction.
Motivated by this, we propose \emph{Conditional Contrast Regularization} (CCR), a training-time regularizer for diffusion RL.
Rather than directly distilling a guided target, CCR uses the frozen reference model's conditional contrast as a local structural prior around the current model prediction.
In this way, CCR complements reward optimization with pretrained quality-related structure, encouraging the learned update to remain compatible with high-quality generation.

Our experiments across multiple reward settings and backbones support this perspective.
Standard RL often exhibits unstable reward--quality trade-offs, whereas CCR more consistently improves visual clarity, prompt alignment, and perceptual quality while maintaining competitive reward.
We further show that generic regularization methods are insufficient from a directional perspective, that simply enabling external CFG during training and inference cannot substitute for CCR, and that backbone-dependent differences in projection dynamics help explain when standard RL succeeds or fails to preserve pretrained quality-related structure.

Our contributions are summarized as follows.
\begin{itemize}
    \item We provide a unified geometric view of CFG and RL by casting both as correction vectors in a common space, and identify the projection of the learned RL correction onto the pretrained CFG correction as a useful diagnostic of quality-related structural preservation.
    \item We show through geometric analysis and rotation intervention that the pretrained CFG direction carries useful quality-related information, while reward-driven RL updates do not necessarily preserve it.
    \item We propose Conditional Contrast Regularization (CCR), a training-time regularizer for diffusion RL that uses the frozen reference model's conditional contrast as a local structural prior, improving the reward--quality trade-off across multiple reward settings and backbones.
\end{itemize}


\section{Related Work}

\paragraph{RL for diffusion models.}
Recent studies have explored RL for post-training diffusion and flow-based generative models under preference, alignment, or task-specific rewards~\cite{black2023training,lee2023aligning,fan2023dpok,wallace2024diffusion,yang2024using,li2025divergence,yuan2024self,liu2025flow,xue2025dancegrpo,zheng2025diffusionnft,wang2026gdro,zhang2026drm}.
These methods show that reward optimization can substantially improve target metrics, but it does not necessarily preserve holistic generation quality, often leading to unstable trade-offs across visual quality, detail, and prompt alignment~\cite{gao2023scaling,dandapanthula2026we,mao2025image,wang2026gdro}.
Existing attempts to mitigate this issue mainly follow two directions: regularizing the policy toward a reference distribution, e.g., through KL-style or data-based regularization~\cite{fan2023dpok,liu2025flow,ye2025data,he2025gardo}, or improving the reward side through stronger reward models or better reward design~\cite{liu2026beyond,zhao2026marble,zhang2026drm}.
Our work differs from both lines by introducing a CFG-inspired structural prior during RL, aiming to preserve quality-related structure that standard reward optimization does not reliably maintain.

\paragraph{CFG and CFG distillation.}
CFG is the standard mechanism for improving conditional generation in diffusion models at inference time~\cite{ho2022classifier,nichol2021glide}.
Recent work has studied both how to improve CFG itself and how to reduce or remove explicit CFG at inference time through guidance distillation or calibration~\cite{chung2024cfg++,sadat2024eliminating,ahn2024self,li2025towards,sadat2025guidance,fan2025cfg,sadat2024no,meng2023distillation,yin2024one,jensen2025efficient,chen2025visual,tang2025diffusion,zheng2025direct,li2026mclr}.
Recent analysis further suggests that, in DMD-style methods, the injected CFG signal itself may be a major source of performance gain~\cite{liu2025decoupled}, motivating the use of CFG-related structure as a prior during RL optimization.
In contrast to these works, we repurpose the frozen reference model's conditional contrast as a training signal, rather than using CFG at inference time or directly distilling guided predictions.

\section{Geometric Analysis}
\label{sec:geometric_analysis}

\begin{figure}[t]
    \centering
    \includegraphics[width=0.95\columnwidth]{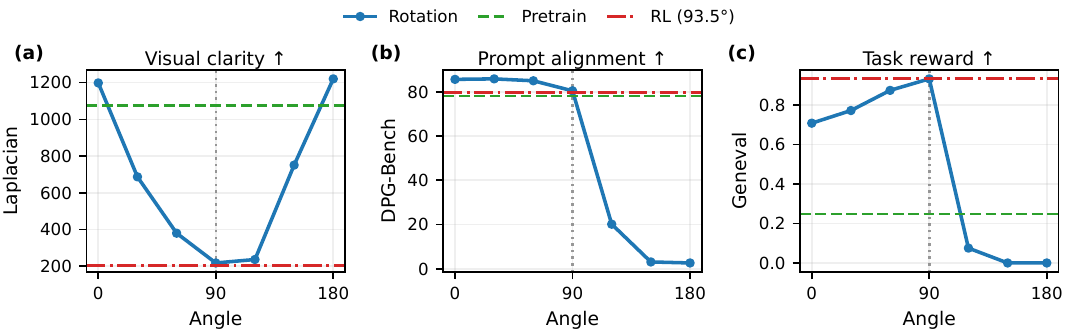}
    \caption{
    Effect of rotating the learned correction after RL with GenEval reward.
    (a) Visual clarity. (b) Prompt alignment. (c) GenEval reward.
    Dashed lines mark pretrained and RL baselines.
    }
    \label{fig:rotation_intervention}
\end{figure}

\begin{figure}[t]
    \centering
    \includegraphics[width=0.8\columnwidth]{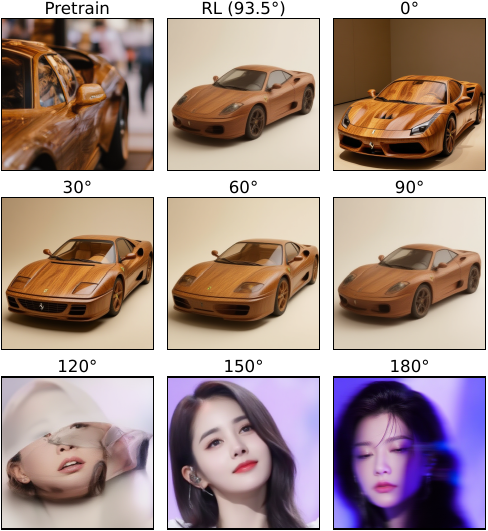}
    \caption{
    Visualization of rotating the learned correction. 
    All images use the prompt "a Ferari car that is made out of wood".
    }
    \label{fig:rotate_exp_samples}
\end{figure}

\subsection{A Unified View of CFG and RL as Corrections}

To understand the reward--quality mismatch in diffusion RL, we adopt a unified geometric view in which the effect of guidance is represented as a correction relative to a baseline prediction.
Under this view, the inference-time effects of CFG and RL can both be written as vectors in the same prediction space, allowing their relationship to be analyzed directly within a common denoising geometry.

Let $v_{\theta_{\mathrm{pre}}}(x_t,c,t)$ denote the pretrained model's conditional prediction at noisy state $x_t$, condition $c$, and time $t$~\cite{salimans2022progressive,lipman2022flow}.
For CFG, we define the correction applied by turning CFG on relative to the pretrained conditional prediction as
\begin{equation}
\begin{aligned}
\Delta_{\mathrm{cfg}}(x_t,c,t)
&=
v_{\theta_{\mathrm{pre}}}^{\mathrm{cfg}}(x_t,c,t)-v_{\theta_{\mathrm{pre}}}(x_t,c,t) \\
&\propto
v_{\theta_{\mathrm{pre}}}(x_t,c,t)-v_{\theta_{\mathrm{pre}}}(x_t,\varnothing,t).
\end{aligned}
\end{equation}

Here, \(\varnothing\) denotes the empty condition.
From this perspective, CFG can be interpreted as a conditional contrast correction.

Similarly, for the model after RL, we define its learned correction relative to the pretrained conditional prediction as
\begin{equation}
\Delta_{\mathrm{learn}}(x_t,c,t)
=
v_{\theta_{\mathrm{rl}}}(x_t,c,t)-v_{\theta_{\mathrm{pre}}}(x_t,c,t).
\end{equation}

In this view, the key question becomes how the learned correction \(\Delta_{\mathrm{learn}}\) relates to the CFG correction \(\Delta_{\mathrm{cfg}}\), in particular whether the former aligns with, deviates from, or opposes the latter.

\subsection{A Diagnostic of Structural Preservation}

Within the unified view above, we seek a simple trajectory-level quantity that summarizes the relationship between the correction learned by RL and the pretrained CFG correction.
To this end, we define the stepwise projection coefficient
\begin{equation}
\alpha(x_t,c,t)
=
\frac{
\left\langle
\Delta_{\mathrm{learn}}(x_t,c,t),
\Delta_{\mathrm{cfg}}(x_t,c,t)
\right\rangle
}{
\|\Delta_{\mathrm{cfg}}(x_t,c,t)\|^2
}.
\end{equation}

Since final generation depends on the full denoising trajectory rather than a single step, we further define the average projection coefficient as
\begin{equation}
\bar{\alpha}(c)
=
\frac{1}{|\mathcal{T}|}\sum_{t\in\mathcal{T}}\alpha(x_t,c,t),
\end{equation}
where $\mathcal{T}$ denotes the set of denoising timesteps.

This quantity admits a simple local first-order interpretation.
Since CFG applies a correction around the pretrained conditional prediction that is associated with improved generation quality, we use $\Delta_{\mathrm{cfg}}$ as a local proxy for a quality-related direction.
The signed projection of $\Delta_{\mathrm{learn}}$ onto this direction then provides a geometric summary of whether the learned correction preserves or opposes those benefits.

This diagnostic is especially relevant in our setting.
Standard RL can substantially improve target reward, yet broader generation quality is not always preserved.
If some of this reward--quality mismatch arises because RL fails to retain a beneficial CFG-related component, then the average projection coefficient provides a useful way to characterize that structural drift over the denoising trajectory.

\subsection{Rotation Intervention on Learned Correction}

\begin{figure*}[t]
    \centering
    \includegraphics[width=0.77\textwidth]{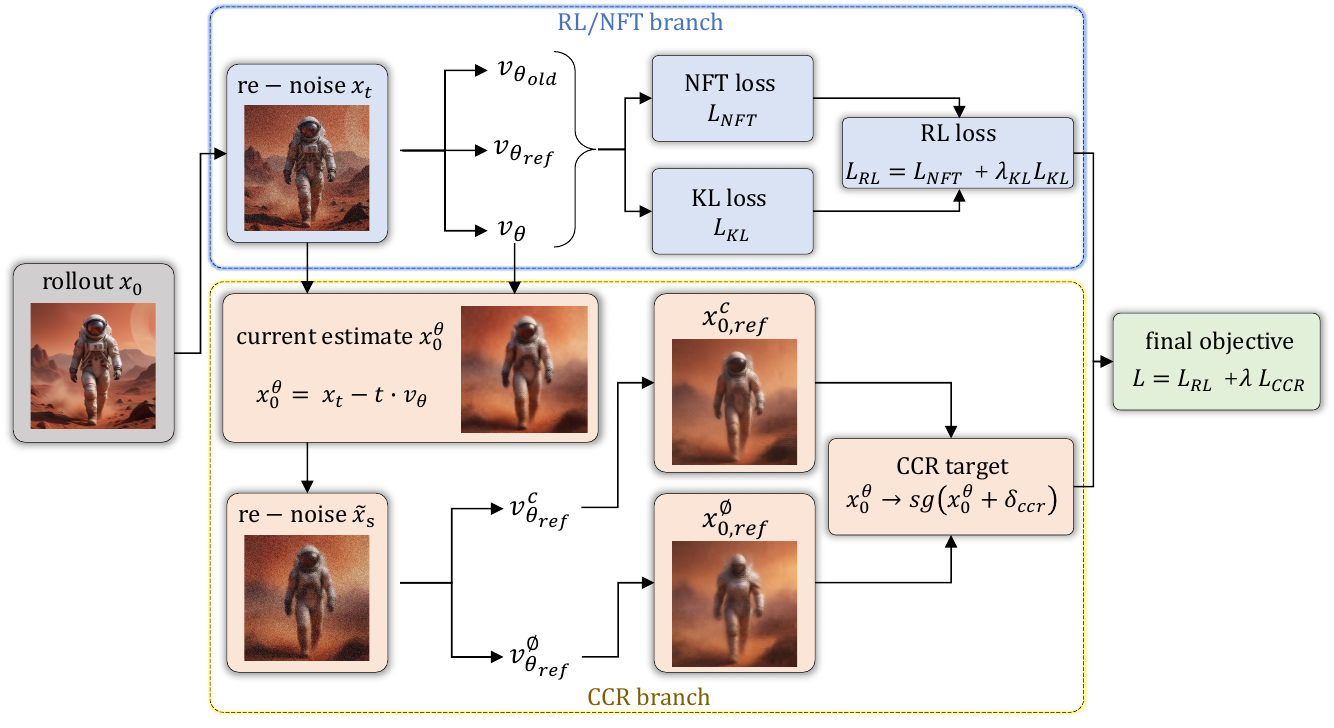}
    \caption{Method overview. From a rollout sample $x_0$, the RL/NFT branch computes $\mathcal{L}_{\mathrm{RL}}=\mathcal{L}_{\mathrm{NFT}}+\lambda_{\mathrm{KL}}\mathcal{L}_{\mathrm{KL}}$, while the CCR branch builds a local contrast target from conditional and unconditional reference predictions. The final objective is $\mathcal{L}=\mathcal{L}_{\mathrm{RL}}+\lambda\mathcal{L}_{\mathrm{CCR}}$.}
    \label{fig:training_overview}
\end{figure*}

To test more directly whether the pretrained CFG correction contains useful quality-related information, we design a rotation intervention that changes the direction of the learned correction relative to $\Delta_{\mathrm{cfg}}$ during sampling.
This intervention is constructed from the decomposition of the projection coefficient itself.

From the definition of the projection coefficient, we can write
\begin{equation}
\alpha(x_t,c,t)
=
\frac{
\langle \Delta_{\mathrm{learn}},\Delta_{\mathrm{cfg}}\rangle
}{
\|\Delta_{\mathrm{cfg}}\|^2
}
=
\frac{\|\Delta_{\mathrm{learn}}\|}{\|\Delta_{\mathrm{cfg}}\|}\cos\phi(x_t,c,t),
\end{equation}
where $\phi(x_t,c,t)$ is the angle between $\Delta_{\mathrm{learn}}$ and $\Delta_{\mathrm{cfg}}$.
This decomposition suggests a simple intervention that preserves the magnitude of the learned correction while controlling its direction relative to the pretrained CFG correction.

Based on this idea, at each denoising step we construct the following intervention. Let
\begin{equation}
e_{\parallel}
=
\frac{\Delta_{\mathrm{cfg}}}{\|\Delta_{\mathrm{cfg}}\|}
\end{equation}
be the unit vector along the pretrained CFG correction direction, and let $e_{\perp}$ be a unit vector constructed from the component of $\Delta_{\mathrm{learn}}$ orthogonal to $\Delta_{\mathrm{cfg}}$.
For a prescribed angle $\phi$, we define the rotated correction as
\begin{equation}
\Delta_{\mathrm{rot}}(\phi)
=
\|\Delta_{\mathrm{learn}}\|
\left(
\cos\phi\, e_{\parallel}
+
\sin\phi\, e_{\perp}
\right).
\end{equation}

In practice, we apply a single fixed angle $\phi$ to all denoising steps in a sampling trajectory, ranging from $0^\circ$ to $180^\circ$.
Here, $0^\circ$ corresponds to full alignment with the pretrained CFG correction, $90^\circ$ corresponds to orthogonality, and angles larger than $90^\circ$ place the correction in the anti-CFG regime.
For reference, the RL correction lies at an average angle of roughly $93.5^\circ$ to the pretrained CFG correction.

As shown in Figures~\ref{fig:rotation_intervention} and~\ref{fig:rotate_exp_samples}, the results display a clear angle-dependent pattern.
Rotating the learned correction toward the pretrained CFG correction improves visual clarity and prompt alignment, whereas rotating it into the anti-CFG regime rapidly degrades these attributes and often leads to semantic errors or generation failure.
At the same time, the original RL correction typically achieves the highest target reward, and rotating it toward any tested alternative direction reduces reward.

These findings support two conclusions.
First, the pretrained CFG correction indeed carries useful quality-related information, and the average projection coefficient captures a meaningful aspect of whether RL preserves that structure.
Second, the reward-optimal direction learned by standard RL is not necessarily aligned with the direction that best preserves broader generation quality.
This suggests that the relevant structural information should be incorporated during training, rather than imposed only as an intervention at inference time.

\begin{table*}[t]
\centering
\small
\setlength{\tabcolsep}{2.0pt}
\caption{Effect of inference-time CFG scale on pretrained Z-Image.
The best and second-best results are shown in
\textbf{bold} and \underline{underlined}.}
\label{tab:cfg_scale}
\begin{tabular*}{\textwidth}{@{\extracolsep{\fill}}l l c c c c c c c c c c c}
\toprule
\multirow{2}{*}{\textbf{Setting}}
& \multirow{2}{*}{\textbf{Method}}
& \multicolumn{2}{c}{\textbf{Visual Clarity $\uparrow$}}
& \multicolumn{2}{c}{\textbf{Prompt Alignment $\uparrow$}}
& \multicolumn{3}{c}{\textbf{Perceptual Quality $\uparrow$}}
& \multicolumn{4}{c}{\textbf{Reward $\uparrow$}} \\
\cmidrule(lr){3-4}
\cmidrule(lr){5-6}
\cmidrule(lr){7-9}
\cmidrule(lr){10-13}
& & Lap. & CLIP-IQA & DPG & VQA & MUSIQ & MANIQA & DEQA
& GenEval & PickScore & HPSv2.1 & Multi-rew. \\
\midrule
\multirow{3}{*}{Pretrained}
& w/o CFG
& 1064 & 0.62 & 78.5 & 0.69 & 63.2 & 0.35 & 3.02
& 0.258 & 0.713 & 0.218 & 1.157 \\
& w/ CFG=2
& \underline{1192} & \underline{0.78}
& \underline{84.7} & \underline{0.89}
& \underline{67.7} & \underline{0.44} & \underline{3.50}
& \underline{0.558} & \underline{0.754}
& \underline{0.260} & \underline{1.270} \\
& w/ CFG=5
& \textbf{1279} & \textbf{0.88}
& \textbf{86.0} & \textbf{0.93}
& \textbf{70.9} & \textbf{0.53} & \textbf{3.92}
& \textbf{0.667} & \textbf{0.770}
& \textbf{0.279} & \textbf{1.318} \\
\bottomrule
\end{tabular*}
\end{table*}

\begin{table*}[t]
\centering
\small
\caption{Quantitative comparison of RL regularization methods on Z-Image across four reward settings.
CCR achieves a better overall reward–quality trade-off, substantially improving image quality while maintaining competitive reward scores.}
\label{tab:sharpness_metrics}
\begin{tabular*}{0.95\textwidth}{@{\extracolsep{\fill}}l l c c c c c c c c}
\toprule
\multirow{2}{*}{\textbf{Setting}}
& \multirow{2}{*}{\textbf{Method}}
& \multicolumn{2}{c}{\textbf{Visual Clarity $\uparrow$}}
& \multicolumn{2}{c}{\textbf{Prompt Alignment $\uparrow$}}
& \multicolumn{3}{c}{\textbf{Perceptual Quality $\uparrow$}}
& \multirow{2}{*}{\textbf{Reward $\uparrow$}} \\
\cmidrule(lr){3-4}
\cmidrule(lr){5-6}
\cmidrule(lr){7-9}
& & Lap. & CLIP-IQA & DPG & VQA & MUSIQ & MANIQA & DEQA & \\
\midrule

\multirow{5}{*}{GenEval}
& Default KL
& 208 & 0.75 & 79.8 & 0.82 & 67.6 & 0.39 & 3.24 & 0.938 \\
& Larger KL
& 331 & 0.60 & \textbf{83.3} & \textbf{0.87}
& 64.7 & 0.35 & 3.17 & \underline{0.952} \\
& DDRL~\cite{ye2025data}
& 542 & \underline{0.80} & 81.8 & \underline{0.86}
& \underline{70.0} & \underline{0.50} & \underline{3.64} & 0.923 \\
& GARDO~\cite{he2025gardo}
& \underline{821} & 0.73 & 79.3 & 0.78
& 65.5 & 0.49 & 3.36 & 0.758 \\
& \textbf{CCR}
& \textbf{841} & \textbf{0.96} & \underline{83.1} & \textbf{0.87}
& \textbf{74.7} & \textbf{0.67} & \textbf{4.33} & \textbf{0.960} \\
\midrule

\multirow{5}{*}{PickScore}
& Default KL
& 229 & 0.67 & \underline{82.8} & \underline{0.90}
& 65.7 & 0.39 & 3.71 & \textbf{0.842} \\
& Larger KL
& 443 & 0.57 & 82.2 & \underline{0.90}
& 66.1 & 0.35 & 3.47 & \underline{0.830} \\
& DDRL~\cite{ye2025data}
& \underline{732} & \underline{0.84} & \underline{82.8} & 0.89
& \underline{70.0} & \underline{0.51} & \underline{3.77} & 0.814 \\
& GARDO~\cite{he2025gardo}
& 363 & 0.54 & 75.5 & 0.69
& 57.2 & 0.30 & 2.47 & 0.749 \\
& \textbf{CCR}
& \textbf{1374} & \textbf{0.93} & \textbf{85.3} & \textbf{0.93}
& \textbf{73.8} & \textbf{0.67} & \textbf{4.35} & \underline{0.830} \\
\midrule

\multirow{5}{*}{HPSv2.1}
& Default KL
& \underline{928} & \underline{0.80} & 80.4 & 0.82
& \underline{73.6} & 0.45 & \underline{4.05} & \textbf{0.377} \\
& Larger KL
& 875 & 0.53 & 82.8 & 0.86
& 68.5 & 0.35 & 3.38 & 0.350 \\
& DDRL~\cite{ye2025data}
& 805 & \underline{0.80} & \underline{83.5} & \underline{0.89}
& 67.8 & \underline{0.46} & \underline{3.41} & 0.330 \\
& GARDO~\cite{he2025gardo}
& 808 & 0.54 & 76.8 & 0.68
& 62.5 & 0.34 & 2.91 & 0.271 \\
& \textbf{CCR}
& \textbf{1919} & \textbf{0.97} & \textbf{85.3} & \textbf{0.91}
& \textbf{76.6} & \textbf{0.68} & \textbf{4.46} & \underline{0.367} \\
\midrule

\multirow{5}{*}{Multi-rew.}
& Default KL
& 755 & 0.74 & \underline{85.8} & \underline{0.93}
& \underline{72.1} & 0.41 & \underline{3.87} & \textbf{1.456} \\
& Larger KL
& \underline{954} & 0.65 & 85.7 & \underline{0.93}
& 68.7 & 0.38 & 3.51 & 1.407 \\
& DDRL~\cite{ye2025data}
& 736 & \underline{0.85} & 85.1 & 0.92
& 70.2 & \underline{0.50} & 3.74 & 1.384 \\
& GARDO~\cite{he2025gardo}
& 510 & 0.55 & 78.2 & 0.74
& 57.6 & 0.32 & 2.71 & 1.233 \\
& \textbf{CCR}
& \textbf{2063} & \textbf{0.97} & \textbf{86.8} & \textbf{0.94}
& \textbf{75.3} & \textbf{0.69} & \textbf{4.36} & \underline{1.431} \\
\bottomrule
\end{tabular*}
\end{table*}


\section{Method}

The geometric analysis above suggests that the key challenge in diffusion RL is not merely to limit policy drift, but to preserve a beneficial component of the learned update that is aligned with pretrained CFG-related structure.
Based on this insight, we propose Conditional Contrast Regularization (CCR), a training-time regularizer for diffusion RL.
Our framework follows a DiffusionNFT-style RL setup~\cite{zheng2025diffusionnft}.
As illustrated in Figure~\ref{fig:training_overview}, the RL/NFT branch optimizes target reward, while the CCR branch regularizes the update direction using pretrained CFG-related structure.

\subsection{RL Framework}

Let $v_{\theta}(x_t, c, t)$ denote the velocity prediction under the current policy, and let $v_{\theta_{\mathrm{old}}}(x_t, c, t)$ and $v_{\theta_{\mathrm{ref}}}(x_t, c, t)$ denote the corresponding old-policy and frozen reference predictions.
Given a clean latent $x_0$ collected from rollout, a sampled time $t \in [0,1]$, and Gaussian noise $\epsilon \sim \mathcal{N}(0,I)$, we form
\begin{equation}
x_t = (1-t)x_0 + t\epsilon,
\qquad
v = \epsilon - x_0.
\end{equation}

Following DiffusionNFT, we use implicit positive and negative branches
\begin{equation}
v_{\theta}^{+}
=
(1-\beta)v_{\theta_{\mathrm{old}}}
+
\beta v_{\theta},
\quad
v_{\theta}^{-}
=
(1+\beta)v_{\theta_{\mathrm{old}}}
-
\beta v_{\theta},
\end{equation}
and define
\begin{equation}
\begin{aligned}
\mathcal{L}_{\mathrm{NFT}}
=
\mathbb{E}_{x_0,c,t,\epsilon}
\Big[
&\,r\,\|v_{\theta}^{+}(x_t,c,t)-v\|_2^2 \\
&+(1-r)\,\|v_{\theta}^{-}(x_t,c,t)-v\|_2^2
\Big].
\end{aligned}
\end{equation}
where $r\in[0,1]$ denotes the normalized advantage weight.

The full RL branch loss is
\begin{equation}
\mathcal{L}_{\mathrm{RL}}
=
\mathcal{L}_{\mathrm{NFT}}
+
\lambda_{\mathrm{KL}}\mathcal{L}_{\mathrm{KL}},
\end{equation}
where $\mathcal{L}_{\mathrm{KL}}$ denotes the KL regularization term relative to the reference model~\cite{liu2025flow}.

\subsection{Conditional Contrast Regularization}

The core idea of CCR is to regularize the current policy using a \emph{local conditional contrast} induced by the frozen reference model.
Rather than directly distilling a guided target at the original noisy training state $(x_t,t)$, CCR first follows the current policy's prediction to its clean estimate $x_0^\theta$, and then constructs a local conditional contrast in a re-noised neighborhood around that estimate.
This design allows the regularizer to adapt to the model's current RL-updated prediction while still anchoring it to a pretrained CFG-related direction, so that it acts as a local correction along the current optimization trajectory rather than a fixed target that competes with RL.

Starting from the noisy state $x_t$, we first obtain the current policy's clean estimate
\begin{equation}
x_0^\theta = x_t - t\,v_\theta(x_t,c,t).
\end{equation}

We then sample a re-noising level $s \in [0,1]$ and Gaussian noise $\epsilon' \sim \mathcal{N}(0,I)$, and form
\begin{equation}
\tilde{x}_s=(1-s)x_0^\theta+s\epsilon'.
\end{equation}

At this local re-noised state, we query the frozen reference model under both the conditional input $c$ and the empty condition $\varnothing$, and convert the resulting predictions back to $x$-space:
\begin{equation}
x_{0,\mathrm{ref}}^c
=
\tilde{x}_s-s\,v_{\theta_{\mathrm{ref}}}(\tilde{x}_s,c,s),\,
x_{0,\mathrm{ref}}^\varnothing
=
\tilde{x}_s-s\,v_{\theta_{\mathrm{ref}}}(\tilde{x}_s,\varnothing,s).
\end{equation}

Their difference defines the local reference conditional contrast
\begin{equation}
\Delta_{\mathrm{ref}}^x
=
x_{0,\mathrm{ref}}^c-x_{0,\mathrm{ref}}^\varnothing.
\end{equation}

This conditional contrast is constructed following the conditional--unconditional decomposition used in CFG. It serves as a local approximation to the pretrained conditional contrast structure around the clean state predicted by the current policy.

Using $\Delta_{\mathrm{ref}}^x$, we construct a normalized correction as
\begin{equation}
\delta_{\mathrm{ccr}} = \operatorname{Normalize}\!\left(\Delta_{\mathrm{ref}}^x\right),
\end{equation}
where normalization is applied per sample to control correction magnitude. We then define the stop-gradient target
\begin{equation}
\tilde{x}_0^{\mathrm{ccr}}
=
\operatorname{sg}\!\left(x_0^\theta+\delta_{\mathrm{ccr}}\right),
\end{equation}
and the CCR loss
\begin{equation}
\mathcal{L}_{\mathrm{CCR}}
=
\left\|x_0^\theta-\tilde{x}_0^{\mathrm{ccr}}\right\|_2^2.
\end{equation}

Because the target is stop-gradient, minimizing $\mathcal{L}_{\mathrm{CCR}}$ encourages the current prediction $x_0^\theta$ to move along the local conditional contrast direction induced by the frozen reference model.
In this sense, CCR itself does not attempt to maximize reward, nor does it explicitly optimize the average projection coefficient.
Instead, it acts as a local structural prior that encourages reward-driven updates to retain a beneficial component aligned with CFG-related structure.

\subsection{Final Objective}

Our final training objective is
\begin{equation}
\mathcal{L}
=
\mathcal{L}_{\mathrm{RL}}
+
\lambda \mathcal{L}_{\mathrm{CCR}},
\end{equation}
where $\lambda$ controls the strength of the CCR regularization.
In this way, CCR uses pretrained conditional contrast as a structural prior during training, without overly constraining reward optimization.

\begin{figure}[t]
\centering
\includegraphics[width=0.95\columnwidth]{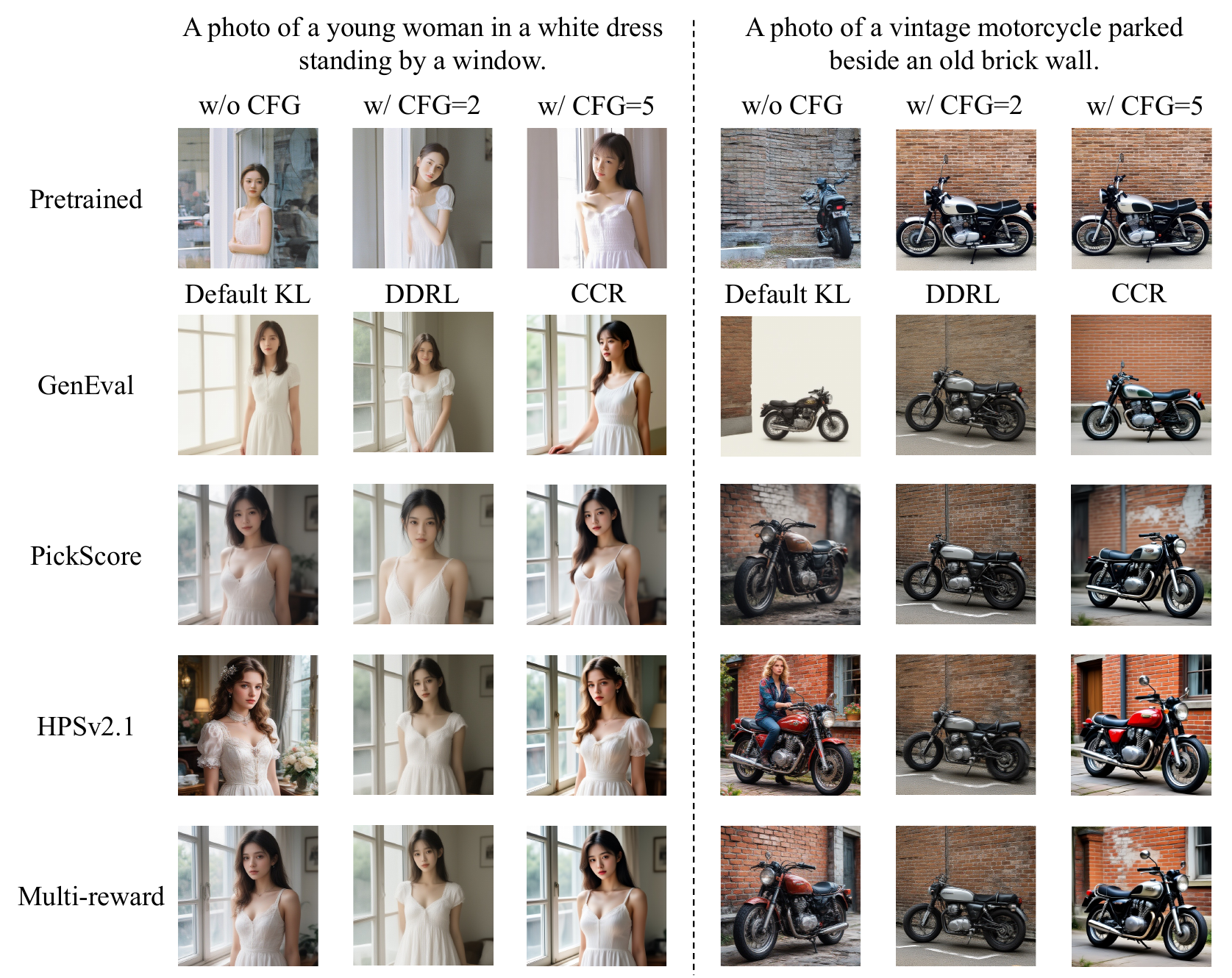}
\caption{
Visual comparison on Z-Image. For each prompt, we show pretrained outputs at different CFG scales and RL outputs from Default KL, DDRL, and CCR under four reward settings.
}
\label{fig:samples}
\end{figure}

\begin{figure}[t]
\centering
\includegraphics[width=0.95\columnwidth]{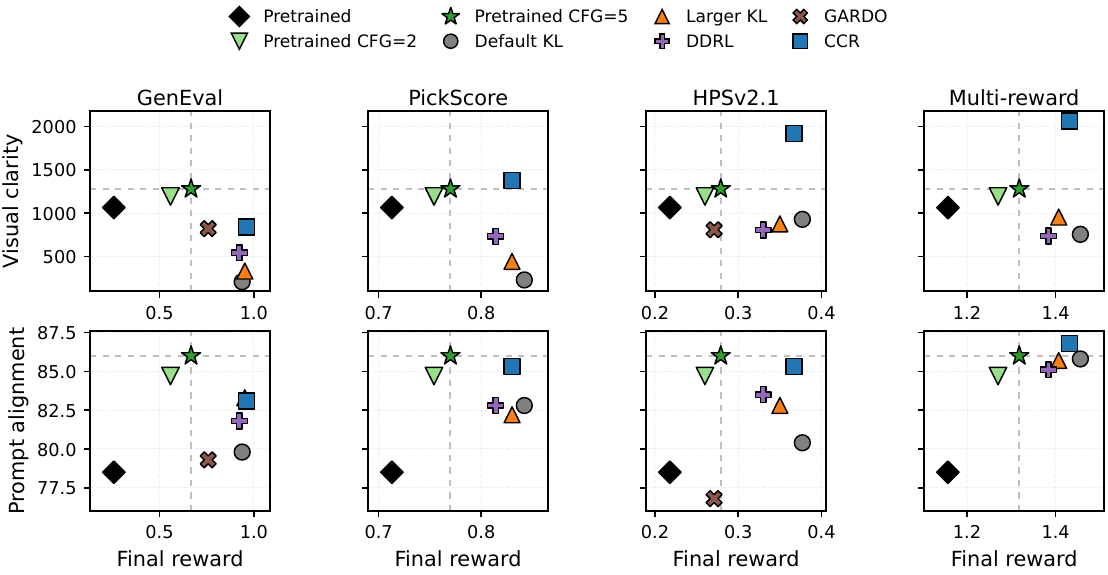}
\caption{
Visual clarity and prompt alignment versus final reward across four reward settings. The top and bottom rows show Laplacian variance and DPG score, respectively.
}
\label{fig:reward_vs_quality}
\end{figure}


\section{Experiments}

We evaluate our method along five dimensions.
First, we compare CCR with several representative regularization baselines for diffusion RL, including default KL regularization, larger KL regularization, data regularization, and dynamic KL regularization, focusing on the resulting reward--quality trade-off.
Second, we examine whether CCR better preserves CFG-related structure during RL by tracking the projection of the learned correction onto the pretrained CFG correction over training. 
Third, we provide additional evidence by comparing with external CFG during training and inference, and by analyzing behavior across different backbones. 
Fourth, we ablate the local structural design of CCR against direct CFG-style distillation. 
Finally, we study the sensitivity to the CCR regularization weight.

\subsection{Experimental Setup}

\paragraph{Models and resolution.}
We conduct our main experiments on the Z-Image base model~\cite{cai2025z} and additionally include Qwen-Image~\cite{wu2025qwen} and SD3.5-Medium~\cite{esser2024scaling} for backbone comparison.
All experiments use $512 \times 512$ resolution.
RL uses LoRA adapters~\cite{hu2022lora} with $\alpha = 64$ and rank $r = 32$.

\paragraph{Reward settings and baselines.}
We consider GenEval~\cite{ghosh2023geneval}, PickScore~\cite{kirstain2023pick}, HPSv2.1~\cite{wu2023human}, and a multi-reward setting combining PickScore, HPSv2.1, and CLIPScore~\cite{hessel2021clipscore}.
For GenEval, we use the same training prompts as FlowGRPO~\cite{liu2025flow}; for the other settings, we use Pick-a-Pic~\cite{kirstain2023pick} prompts.
We compare CCR against four regularization baselines: KL regularization with the default coefficient $\lambda_{\mathrm{KL}}=10^{-4}$, KL regularization with a larger coefficient $\lambda_{\mathrm{KL}}=10^{-2}$, DDRL~\cite{ye2025data} data regularization using images generated by Qwen-Image-2~\cite{zhao2026qwen}, and GARDO~\cite{he2025gardo} dynamic KL regularization.
KL sweeps showed no consistent improvements from increasing KL strength, so we report only the default and larger settings.

\paragraph{Evaluation metrics.}
We evaluate generation quality from three complementary aspects.
For visual clarity, we report variance of Laplacian~\cite{pertuz2013analysis} and CLIP-IQA sharpness~\cite{wang2023exploring}.
For prompt alignment, we report DPG-Bench~\cite{hu2024ella} and VQAScore~\cite{lin2024evaluating}.
For perceptual quality, we report MUSIQ~\cite{ke2021musiq}, MANIQA~\cite{yang2022maniqa}, and DeQA-Score~\cite{you2025teaching} (with Qwen2.5-VL-7B~\cite{Bai2025Qwen25VLTR}).
For each reward setting, we also report the benchmark score.
For geometric analysis, we report the average projection coefficient of the learned RL correction onto the pretrained CFG correction.

\subsection{CCR Improves the Reward--Quality Trade-off}

\begin{figure}[t]
    \centering
    \includegraphics[width=0.95\columnwidth]{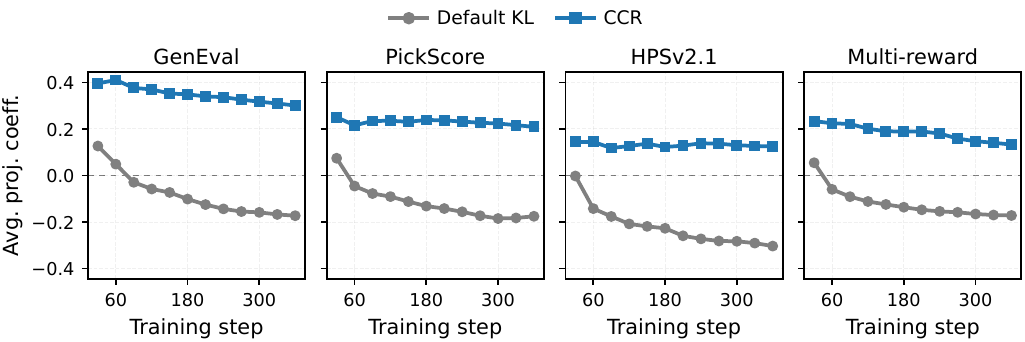}
    \caption{
    Average projection coefficient along the CFG correction during training across four reward settings.
    }
    \label{fig:projection_over_steps}
\end{figure}

\begin{table}[t]
\centering
\small
\setlength{\tabcolsep}{3.0pt}
\caption{
Quantitative evaluation of CCR on Qwen-Image.
}
\label{tab:geneval_small_comparison}
\begin{tabular*}{\columnwidth}{@{\extracolsep{\fill}} l l c c c c @{}}
\toprule
Setting & Method & Lap. $\uparrow$ & DPG $\uparrow$ & MUSIQ $\uparrow$ & GenEval $\uparrow$ \\
\midrule
Pretrained & w/o CFG        & \underline{242} & 85.0 & \underline{68.2} & 0.677 \\
GenEval    & Default KL     & 169 & \underline{85.5} & 66.0 & \textbf{0.947} \\
GenEval    & \textbf{CCR}   & \textbf{744} & \textbf{86.3} & \textbf{72.3} & \underline{0.933} \\
\bottomrule
\end{tabular*}
\end{table}

We first compare CCR against pretrained models and several representative RL regularization baselines on Z-Image.
Our goal is not only to test whether a method improves the target reward, but whether it can do so while preserving broader generation quality.
We therefore evaluate both reward and three complementary aspects of quality, namely visual clarity, prompt alignment, and perceptual quality.

As a reference, we begin by examining pretrained Z-Image under different inference-time CFG scales.
As shown in Table~\ref{tab:cfg_scale} and Figure~\ref{fig:samples}, enabling CFG consistently improves multiple aspects of generation quality.
This suggests that the pretrained CFG direction captures useful structure for high-quality conditional generation.

We then compare RL regularization baselines and CCR across reward settings.
As shown in Table~\ref{tab:sharpness_metrics} and Figure~\ref{fig:samples}, different regularization strategies exhibit markedly different reward--quality trade-offs.
Default KL regularization often achieves strong reward optimization, but does not reliably preserve overall generation quality.
Increasing the KL strength does not consistently resolve this issue; in several cases it lowers reward without stably restoring quality.
Data regularization can partially improve certain quality dimensions, but its gains are uneven across settings and metrics.
Dynamic KL regularization is also unstable across settings and can substantially reduce reward.

By contrast, CCR more consistently preserves or improves visual clarity, prompt alignment, and perceptual quality across reward settings, achieving the best or near-best results on almost all quality metrics while maintaining competitive reward.
For GenEval, CCR slightly outperforms the default KL baseline in reward, whereas under PickScore, HPSv2.1, and multi-reward settings, the default KL baseline is sometimes marginally higher.
These reward differences are small relative to the substantial gains achieved by CCR.
As shown in Figure~\ref{fig:reward_vs_quality}, CCR remains close to the default KL baseline along the reward axis while markedly shifting upward on the quality axes, leading to a better overall reward--quality trade-off.

\subsection{CCR Preserves CFG Structure During RL}

\begin{figure}[t]
    \centering
    \includegraphics[width=0.95\columnwidth]{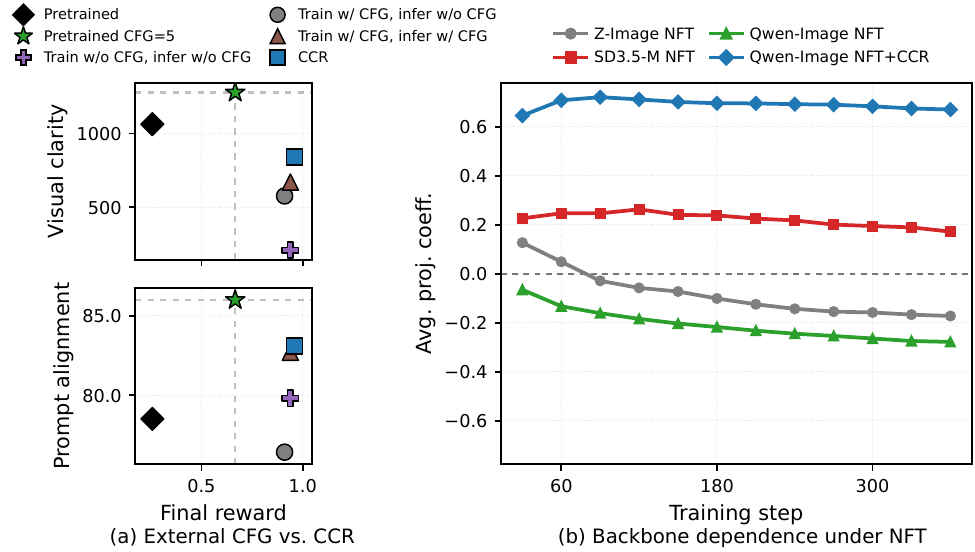}
    \caption{
    Additional analyses. (a) External CFG versus CCR on Z-Image with GenEval reward, showing Laplacian variance (top) and DPG score (bottom) versus final reward. (b) Dynamics of projection coefficients across backbones during NFT training.
    }
    \label{fig:additional_observations}
\end{figure}

We next examine whether the geometric diagnostic introduced in Section~\ref{sec:geometric_analysis} helps explain the reward--quality trade-offs above.
Specifically, we track the average projection coefficient of the learned RL correction onto the pretrained CFG correction throughout training.

Figure~\ref{fig:projection_over_steps} shows that, for NFT optimized with default KL regularization, the average projection coefficient typically decreases during training and often becomes negative at later checkpoints, indicating drift away from the pretrained CFG-related direction.
This provides evidence that reward can continue to improve even as the RL correction drifts away from the pretrained CFG correction, while broader generation quality becomes less reliably preserved.

In contrast, CCR more consistently maintains a positive projection coefficient across reward settings.
This suggests that CCR preserves a structural component aligned with the pretrained CFG correction.
These geometric trends are consistent with the quality results: methods that fail to preserve a positive CFG-aligned component tend to perform worse in visual clarity, prompt alignment, and perceptual quality, whereas CCR maintains stronger quality preservation together with a more stable projection pattern.

\subsection{Additional Evidence from Guided Training and Backbone Comparisons}

\begin{table}[t]
\centering
\small
\setlength{\tabcolsep}{1.0pt}
\caption{Comparison between direct CFG distillation and CCR.}
\label{tab:direct_distill_vs_ccr}
\begin{tabular*}{\columnwidth}{@{\extracolsep{\fill}} l l c c c c c @{}}
\toprule
Setting & Method & Lap. $\uparrow$ & DPG $\uparrow$ & MUSIQ $\uparrow$ & Reward $\uparrow$ & $\bar{\alpha}_{\text{final}}$ \\
\midrule
\multirow{3}{*}{GenEval} & Default KL  & 208 & 79.8 & 67.6 & 0.938 & -0.173 \\
& CFG distill. & \textbf{1743} & 81.5 & 73.1 & 0.935 & 0.606 \\
& \textbf{CCR} & 841 & \textbf{83.1} & \textbf{74.7} & \textbf{0.960} & 0.301 \\
\bottomrule
\end{tabular*}
\end{table}

\begin{table}[t]
\centering
\small
\setlength{\tabcolsep}{3.0pt}
\caption{
Effect of the CCR weight \(\lambda\) across four reward settings.
}
\label{tab:lambda_all}
\begin{tabular*}{\columnwidth}{@{\extracolsep{\fill}} l l c c c c @{}}
\toprule
Setting & Method & Lap. $\uparrow$ & DPG $\uparrow$ & Reward $\uparrow$ & $\bar{\alpha}_{\text{final}}$ \\
\midrule
\multirow{4}{*}{GenEval}
& Default KL               & 208  & 79.8 & 0.938 & -0.173 \\
& CCR ($\lambda$=0.03)     & 623  & \textbf{83.4} & 0.950 & 0.112 \\
& CCR ($\lambda$=0.05)     & 841  & 83.1 & \textbf{0.960} & 0.301 \\
& CCR ($\lambda$=0.10)     & \textbf{1487} & 80.9 & 0.944 & 0.213 \\
\midrule

\multirow{4}{*}{PickScore}
& Default KL               & 229  & 82.8 & \textbf{0.842} & -0.175 \\
& CCR ($\lambda$=0.03)     & 535  & 84.3 & 0.839 & 0.077 \\
& CCR ($\lambda$=0.05)     & 1374 & \textbf{85.3} & 0.830 & 0.209 \\
& CCR ($\lambda$=0.10)     & \textbf{1558} & 84.3 & 0.827 & 0.231 \\
\midrule

\multirow{4}{*}{HPSv2.1}
& Default KL               & 928  & 80.4 & \textbf{0.377} & -0.303 \\
& CCR ($\lambda$=0.03)     & 1507 & 85.1 & 0.373 & -0.029 \\
& CCR ($\lambda$=0.05)     & 1919 & 85.3 & 0.367 & 0.124 \\
& CCR ($\lambda$=0.10)     & \textbf{2050} & \textbf{86.1} & 0.357 & 0.105 \\
\midrule

\multirow{4}{*}{Multi-rew.}
& Default KL               & 755  & 85.8 & \textbf{1.456} & -0.172 \\
& CCR ($\lambda$=0.03)     & 1392 & 86.8 & 1.448 & 0.018 \\
& CCR ($\lambda$=0.05)     & \textbf{2063} & \textbf{86.8} & 1.431 & 0.133 \\
& CCR ($\lambda$=0.10)     & 1889 & 76.2 & 1.385 & 0.237 \\
\bottomrule
\end{tabular*}
\end{table}

Figure~\ref{fig:additional_observations}(a) compares CCR with a baseline that enables external CFG during both training and inference on GenEval using Z-Image.
Although this guided baseline partially improves visual clarity and prompt alignment, it is still generally weaker than CCR.
This suggests that the benefit of CCR does not come from injecting external guidance during training or inference, but from regularizing the RL optimization trajectory with a pretrained structural prior.

We further compare standard RL across Z-Image, Qwen-Image, and SD3.5-Medium.
On Qwen-Image, as shown in Table~\ref{tab:geneval_small_comparison}, we observe a clear reward--quality mismatch: standard RL substantially improves reward and some related metrics, but weakens visual clarity and perceptual quality.
CCR mitigates this mismatch and restores a stronger quality profile, while its projection dynamics remain consistent with the CFG-related structural interpretation, as shown in Figure~\ref{fig:additional_observations}(b).
By contrast, on SD3.5-Medium the projection coefficient stays positive throughout training under RL with default KL regularization, and the quality gap relative to CCR is correspondingly smaller.
These results suggest that the benefit of CCR is backbone dependent: it is especially useful when standard RL fails to preserve a CFG-aligned structural component.
This observation is consistent with DiffusionNFT~\cite{zheng2025diffusionnft}, which reports RL on SD3.5-Medium can achieve high-quality CFG-free generation.

\subsection{Ablation of the Local Contrast Construction}

We ablate the re-noising step in CCR to examine whether the improvement comes merely from adding a reference conditional contrast term, or specifically from constructing that contrast around the current model-induced prediction.
Recall that in CCR, the current model first predicts $x_0^\theta$, after which we form a local re-noised state $\tilde{x}_s$ and compute the frozen reference model's conditional contrast there.
This design makes the structural correction depend on the local neighborhood reached by the current model during training.

In the variant that directly distills CFG, the reference conditional contrast is applied directly at the original re-noised state $x_t$.
As shown in Table~\ref{tab:direct_distill_vs_ccr}, this variant produces a larger average projection coefficient, suggesting a more direct competition between the structural regularizer and reward optimization.
While it improves some quality metrics, its reward--quality trade-off is worse than that of CCR.
This indicates that the key is not stronger imitation of guided prediction, but local and adaptive structural regularization around the current RL trajectory.

\subsection{Hyperparameter Sensitivity}

Table~\ref{tab:lambda_all} shows the sensitivity of CCR to the regularization weight $\lambda$.
Across reward settings, moderately larger weights generally improve visual clarity and prompt alignment, while overly strong regularization can be counterproductive and reduce reward or hurt alignment.
Moreover, these results suggest that the average projection coefficient is better viewed as a diagnostic than a quantity to maximize.


\section{Conclusion}

We study reward--quality mismatch, a form of reward hacking, in RL post-training of diffusion models.
Our central argument is that the problem is not merely excessive policy drift, but the failure to reliably preserve the quality-related structure that supports high-quality generation.
By developing a unified geometric view of CFG and RL, we identify pretrained CFG-related structure as a useful prior and propose Conditional Contrast Regularization (CCR) to inject this prior during RL.
Across multiple reward settings and backbones, CCR improves the reward--quality trade-off by more consistently preserving generation quality while maintaining competitive reward.


\section*{LLM Usage Statement}

We used ChatGPT (OpenAI) as an auxiliary tool in preparing this work.
Specifically, it was used to assist with language polishing of the manuscript, improving the presentation of figures, and revising experimental code.
It was not used to generate data, fabricate results, design the core method, or make scientific judgments.
All experiments, data, analyses, and conclusions were conducted and verified by the authors. Any LLM-assisted edits to text, figure presentation, or code were carefully reviewed, validated, and corrected where necessary by the authors.
The authors take full responsibility for the content and integrity of this paper.

{
    \small
    \bibliographystyle{ieeenat_fullname}
    \bibliography{main}
}

\appendix
\section*{Appendix}

\section{A Local Interpretation of the Projection-Based Diagnostic}
\label{app:projection_theory}

This appendix provides a compact interpretation of the projection-based diagnostic used in the main text. The purpose is not to define a full theory of generation quality, but to clarify why projection onto the pretrained CFG correction is a meaningful local summary of whether RL preserves or opposes CFG-related quality structure.

\paragraph{A local quality model around the pretrained conditional prediction.}
Fix a noisy state $(x_t,c,t)$ and let
\begin{equation}
v_0 = v_{\theta_{\mathrm{pre}}}(x_t,c,t)
\end{equation}
denote the pretrained conditional prediction. Recall the pretrained CFG correction
\begin{equation}
\Delta_{\mathrm{cfg}}(x_t,c,t)
=
v_{\theta_{\mathrm{pre}}}(x_t,c,t)
-
v_{\theta_{\mathrm{pre}}}(x_t,\varnothing,t),
\end{equation}
and the learned RL correction
\begin{equation}
\Delta_{\mathrm{learn}}(x_t,c,t)
=
v_{\theta_{\mathrm{rl}}}(x_t,c,t)
-
v_{\theta_{\mathrm{pre}}}(x_t,c,t).
\end{equation}

To describe CFG-related quality locally, introduce a scalar functional
\begin{equation}
Q_t(v),
\end{equation}
which should be understood only as a local explanatory model for quality attributes associated with CFG around $v_0$, such as visual clarity and prompt alignment. Since CFG acts by applying an explicit correction along $\Delta_{\mathrm{cfg}}$, we model the local first-order quality direction as
\begin{equation}
\nabla Q_t(v_0)
=
\lambda_t \Delta_{\mathrm{cfg}}(x_t,c,t),
\qquad
\lambda_t>0.
\label{eq:app_q_grad_model}
\end{equation}
This is an interpretation assumption rather than an independently derived fact: it simply encodes that, in this local model, the CFG correction represents the local direction associated with CFG-related benefits.

\paragraph{Why the projection coefficient is meaningful.}
Consider the RL-updated prediction $v_0+\Delta_{\mathrm{learn}}$. A first-order Taylor expansion around $v_0$ gives
\begin{equation}
Q_t(v_0+\Delta_{\mathrm{learn}})
-
Q_t(v_0)
\approx
\left\langle \nabla Q_t(v_0), \Delta_{\mathrm{learn}} \right\rangle.
\end{equation}
Substituting Eq.~\eqref{eq:app_q_grad_model} yields
\begin{equation}
Q_t(v_0+\Delta_{\mathrm{learn}})
-
Q_t(v_0)
\approx
\lambda_t
\left\langle
\Delta_{\mathrm{cfg}}(x_t,c,t),
\Delta_{\mathrm{learn}}(x_t,c,t)
\right\rangle.
\end{equation}

Now define the stepwise projection coefficient
\begin{equation}
\alpha(x_t,c,t)
=
\frac{
\left\langle \Delta_{\mathrm{learn}}(x_t,c,t), \Delta_{\mathrm{cfg}}(x_t,c,t) \right\rangle
}{
\|\Delta_{\mathrm{cfg}}(x_t,c,t)\|^2
}.
\end{equation}
Then
\begin{equation}
Q_t(v_0+\Delta_{\mathrm{learn}})
-
Q_t(v_0)
\approx
\lambda_t
\|\Delta_{\mathrm{cfg}}(x_t,c,t)\|^2
\alpha(x_t,c,t).
\label{eq:app_q_alpha_relation}
\end{equation}

The prefactor $\lambda_t\|\Delta_{\mathrm{cfg}}(x_t,c,t)\|^2$ is a local scale term. What matters most for the diagnostic is the sign of $\alpha(x_t,c,t)$:
\begin{itemize}
    \item if $\alpha(x_t,c,t)>0$, the learned correction preserves a positive component along the local CFG-related quality direction;
    \item if $\alpha(x_t,c,t)<0$, the learned correction locally opposes that direction.
\end{itemize}
Thus, the projection coefficient should be read primarily as a signed local indicator of whether RL preserves or counteracts CFG-related structure, rather than as a complete scalar measure of quality.

\paragraph{From stepwise projection to average projection.}
Generation quality depends on the full denoising trajectory rather than on a single step. If the trajectory-level effect of RL on CFG-related quality is viewed as an accumulation of local stepwise contributions, then Eq.~\eqref{eq:app_q_alpha_relation} suggests the heuristic relation
\begin{equation}
\Delta Q_{\mathrm{traj}}(c)
\approx
\sum_{t\in\mathcal{T}}
\lambda_t
\|\Delta_{\mathrm{cfg}}(x_t,c,t)\|^2
\alpha(x_t,c,t),
\label{eq:app_traj_quality}
\end{equation}
where $\mathcal{T}$ denotes the set of denoising timesteps. This motivates the average projection coefficient
\begin{equation}
\bar{\alpha}(c)
=
\frac{1}{|\mathcal{T}|}\sum_{t\in\mathcal{T}}\alpha(x_t,c,t),
\end{equation}
which serves as a trajectory-level summary of whether RL tends, on average, to preserve or oppose the pretrained CFG correction direction across denoising.

\paragraph{Why appropriately positive projection matters.}
The diagnostic above only says whether the learned correction keeps a signed component along the CFG correction. In practice, however, CFG-related quality may be reflected through multiple proxy attributes, such as visual clarity and prompt alignment, whose local ascent directions need not be identical to $\Delta_{\mathrm{cfg}}$. This means that merely requiring nonnegative projection is often too weak, while maximizing projection indiscriminately is also not the goal.

To make this more explicit, consider two local proxy quality functionals around $v_0$ with gradients
\begin{equation}
g_1 = \nabla Q_t^{(1)}(v_0),
\qquad
g_2 = \nabla Q_t^{(2)}(v_0),
\end{equation}
and suppose both are acutely aligned with the pretrained CFG correction:
\begin{equation}
\angle(g_1,\Delta_{\mathrm{cfg}})=a<90^\circ,
\qquad
\angle(g_2,\Delta_{\mathrm{cfg}})=b<90^\circ.
\end{equation}
Jointly improving the two proxy qualities requires
\begin{equation}
\langle \Delta_{\mathrm{learn}}, g_1\rangle>0,
\qquad
\langle \Delta_{\mathrm{learn}}, g_2\rangle>0.
\end{equation}
These inequalities define the intersection of two local half-spaces and hence a local cone of admissible directions.

To visualize this cone, consider a two-dimensional local cross-section containing the CFG axis and the projected proxy directions. Let $\theta$ denote the angle of $\Delta_{\mathrm{learn}}$ relative to $\Delta_{\mathrm{cfg}}$, and suppose in this cross-section that $g_1$ and $g_2$ lie on opposite sides of the CFG direction, rotated by $+a$ and $-b$, respectively. Then the two positivity constraints become
\begin{equation}
\cos(\theta-a)>0,
\qquad
\cos(\theta+b)>0,
\end{equation}
which imply
\begin{equation}
\theta \in (a-90^\circ,\; 90^\circ-b).
\label{eq:app_proxy_cone}
\end{equation}
This interval is generally a strict subregion of $(-90^\circ,90^\circ)$, which is the full positive-projection regime relative to $\Delta_{\mathrm{cfg}}$. Therefore, positive projection only guarantees that the learned correction lies on the same signed side of the CFG axis, whereas simultaneously improving multiple CFG-related proxy qualities may favor a narrower angular neighborhood around that axis.

This is the sense in which appropriately positive projection is more meaningful than arbitrary projection maximization. Positive projection is useful because it avoids drift into the anti-CFG regime, but jointly improving multiple CFG-related proxy qualities may require a narrower angular neighborhood around the CFG direction. For this reason, the average projection coefficient is better viewed as a geometric diagnostic than as a standalone quantity to maximize.

\paragraph{Interpretation and limitation.}
Taken together, this local picture supports the use of $\bar{\alpha}$ as a geometric diagnostic. It does not by itself determine final generation quality after RL, which also depends on reward optimization, optimization stability, and nonlinear trajectory effects. However, it provides a meaningful summary of whether the learned RL correction retains a positive component along the pretrained CFG direction or instead drifts systematically against it. Accordingly, this section should be read as an interpretation of why the projection coefficient is a meaningful geometric diagnostic, rather than as a proof that it fully characterizes post-RL generation quality.

\section{Rotation Intervention as Support for the Projection Diagnostic}
\label{app:rotation_intervention}

This appendix explains why the rotation intervention in the main text directly supports the projection-based diagnostic. The goal is to test whether the diagnostic is merely correlational, or whether it captures a functionally meaningful aspect of the quality degradation observed after RL.

\paragraph{Projection coefficient as scale times angle.}
Recall
\begin{equation}
\alpha(x_t,c,t)
=
\frac{
\left\langle \Delta_{\mathrm{learn}}(x_t,c,t), \Delta_{\mathrm{cfg}}(x_t,c,t)\right\rangle
}{
\|\Delta_{\mathrm{cfg}}(x_t,c,t)\|^2
}.
\end{equation}
Let $\phi(x_t,c,t)$ be the angle between $\Delta_{\mathrm{learn}}(x_t,c,t)$ and $\Delta_{\mathrm{cfg}}(x_t,c,t)$. Then
\begin{equation}
\alpha(x_t,c,t)
=
\frac{\|\Delta_{\mathrm{learn}}(x_t,c,t)\|}{\|\Delta_{\mathrm{cfg}}(x_t,c,t)\|}
\cos\phi(x_t,c,t).
\label{eq:app_proj_reparam}
\end{equation}
So the projection coefficient decomposes into:
\begin{itemize}
    \item a scale term $\|\Delta_{\mathrm{learn}}\|/\|\Delta_{\mathrm{cfg}}\|$;
    \item an angular term $\cos\phi$.
\end{itemize}
This form makes clear that projection depends jointly on relative magnitude and direction.

\paragraph{The intervention directly manipulates the angular factor.}
At each denoising step, define
\begin{equation}
e_{\parallel}
=
\frac{\Delta_{\mathrm{cfg}}}{\|\Delta_{\mathrm{cfg}}\|},
\end{equation}
and let $e_{\perp}$ be a unit vector constructed from the component of $\Delta_{\mathrm{learn}}$ orthogonal to $\Delta_{\mathrm{cfg}}$. For a prescribed angle $\theta$, define the rotated correction
\begin{equation}
\Delta_{\mathrm{rot}}(\theta)
=
\|\Delta_{\mathrm{learn}}\|
\left(
\cos\theta\,e_{\parallel}
+
\sin\theta\,e_{\perp}
\right).
\label{eq:app_rotated_correction}
\end{equation}
This preserves the norm of the RL correction while replacing its angle relative to the CFG direction by $\theta$.

The projection coefficient of the rotated correction is therefore
\begin{equation}
\alpha_{\mathrm{rot}}(x_t,c,t;\theta)
=
\frac{
\left\langle \Delta_{\mathrm{rot}}(\theta),\Delta_{\mathrm{cfg}}\right\rangle
}{
\|\Delta_{\mathrm{cfg}}\|^2
}
=
\frac{\|\Delta_{\mathrm{learn}}(x_t,c,t)\|}{\|\Delta_{\mathrm{cfg}}(x_t,c,t)\|}
\cos\theta.
\label{eq:app_rot_proj}
\end{equation}
Thus, the intervention keeps the RL-induced scale term and directly controls only the angular term. It is therefore not an arbitrary manipulation, but a direct intervention on the geometric quantity underlying the diagnostic.

\paragraph{Trajectory-level form.}
If the same fixed angle $\theta$ is used across the whole denoising trajectory, then
\begin{equation}
\begin{split}
\bar{\alpha}_{\mathrm{rot}}(c;\theta)
&=
\frac{1}{|\mathcal{T}|}\sum_{t\in\mathcal{T}}
\alpha_{\mathrm{rot}}(x_t,c,t;\theta) \\
&=
\cos\theta\cdot
\frac{1}{|\mathcal{T}|}\sum_{t\in\mathcal{T}}
\frac{\|\Delta_{\mathrm{learn}}(x_t,c,t)\|}
{\|\Delta_{\mathrm{cfg}}(x_t,c,t)\|}.
\end{split}
\end{equation}

So under a fixed-angle intervention, the average projection is modulated by a common angular factor while retaining the step-dependent scale information from the RL model.

\paragraph{Experimental procedure.}
We run this intervention on the Z-Image model after 360-step DiffusionNFT training on GenEval without CCR, using the same setup as in the main text. We use $512\times512$ resolution and 30 denoising steps. For each prompt, we fix the initial latent and apply the same angle
\begin{equation}
\theta \in \{0^\circ,30^\circ,60^\circ,90^\circ,120^\circ,150^\circ,180^\circ\}
\end{equation}
to every denoising step in the trajectory.

This is a non-adaptive intervention: the angle is not tuned per sample or per timestep. Precisely because of that, systematic trends under this intervention provide stronger evidence that the projection-based diagnostic captures a meaningful geometric property.

\paragraph{Interpretation.}
The main text shows that rotating the learned correction toward the CFG direction improves visual clarity and prompt alignment, whereas rotating it into the anti-CFG regime degrades both. At the same time, the original RL direction remains best for the target reward. This is exactly the pattern expected if the projection coefficient captures a meaningful quality-related trade-off: preserving a positive CFG-aligned component helps maintain CFG-related quality attributes, but exact alignment with CFG is not the same as reward-optimal behavior. This is why the projection coefficient is useful as a diagnostic, not as a standalone objective to maximize.

\section{Direct Contrast at the Original Re-noised Training State Versus Local Contrast from a Second Re-noising Step}
\label{app:two_stage_ccr}

This appendix clarifies the key structural difference between CCR and a direct CFG-style distillation regularizer. The difference is not simply that both use the frozen reference model's conditional contrast, but that they apply it at different re-noised training locations.

In both cases, training starts from the same RL state $x_t$, which is obtained by re-noising a rollout sample $x_0$. The direct variant applies the reference contrast directly at this original re-noised training state. CCR instead first follows the current model prediction to a clean estimate, and then performs a second re-noising step to construct a new local state around that prediction. The distinction is therefore between applying the structural correction at the original re-noised training state and applying it at a prediction-conditioned re-noised state.

For clarity, we compare the two regularizers at the level of induced local correction directions, rather than exact parameter-space optimization updates.

\paragraph{Direct contrast at the original re-noised training state.}
Let $v_\theta(x_t,c,t)$ denote the current model prediction at the original RL training state $(x_t,c,t)$, and let
\begin{equation}
\delta v_{\mathrm{RL}}(x_t,c,t)
\end{equation}
denote the update induced by the RL objective at that state. A direct CFG-style regularizer uses the frozen reference contrast computed at the same state:
\begin{equation}
\Delta_{\mathrm{ref}}^{v}(x_t,c,t)
=
v_{\theta_{\mathrm{ref}}}(x_t,c,t)
-
v_{\theta_{\mathrm{ref}}}(x_t,\varnothing,t).
\end{equation}
Formally, the resulting update can be written as
\begin{equation}
v_\theta(x_t,c,t)
\rightarrow
v_\theta(x_t,c,t)
+
\delta v_{\mathrm{RL}}(x_t,c,t)
+
\lambda \Delta_{\mathrm{ref}}^{v}(x_t,c,t),
\label{eq:direct_cfg_update}
\end{equation}
where $\lambda>0$ is the regularization weight. Thus, the extra structural correction is tied directly to the original re-noised training state.

\paragraph{CCR constructs the contrast after following the current prediction.}
CCR starts from the same original training state $x_t$, but does not apply the reference contrast there directly. Instead, it first follows the current prediction to its clean estimate
\begin{equation}
x_0^\theta = x_t - t\,v_\theta(x_t,c,t).
\label{eq:app_x0theta}
\end{equation}
It then performs a second re-noising step:
\begin{equation}
\tilde{x}_s=(1-s)x_0^\theta+s\epsilon',
\qquad
\epsilon'\sim\mathcal N(0,I).
\label{eq:app_xs}
\end{equation}
Substituting Eq.~\eqref{eq:app_x0theta} gives
\begin{equation}
\tilde{x}_s
=
(1-s)\bigl(x_t-t\,v_\theta(x_t,c,t)\bigr)+s\epsilon',
\end{equation}
so the second re-noised state $\tilde{x}_s$ depends on the current prediction itself.

At $(\tilde{x}_s,s)$, the frozen reference model yields
\begin{equation}
\begin{aligned}
x_{0,\mathrm{ref}}^{c}
&=
\tilde{x}_s - s\,v_{\theta_{\mathrm{ref}}}(\tilde{x}_s,c,s), \\
x_{0,\mathrm{ref}}^{\varnothing}
&=
\tilde{x}_s - s\,v_{\theta_{\mathrm{ref}}}(\tilde{x}_s,\varnothing,s).
\end{aligned}
\end{equation}
whose difference defines the local reference contrast
\begin{equation}
\Delta_{\mathrm{ref}}^{x}
=
x_{0,\mathrm{ref}}^{c}-x_{0,\mathrm{ref}}^{\varnothing}
=
-s\Bigl(
v_{\theta_{\mathrm{ref}}}(\tilde{x}_s,c,s)
-
v_{\theta_{\mathrm{ref}}}(\tilde{x}_s,\varnothing,s)
\Bigr).
\label{eq:app_xv_relation}
\end{equation}

Since CCR applies this correction in $x$-space around $x_0^\theta$, the induced update in $v$-space can be written formally as
\begin{equation}
v_\theta(x_t,c,t)
\rightarrow
v_\theta(x_t,c,t)
+
\delta v_{\mathrm{RL}}(x_t,c,t)
+
\delta v_{\mathrm{CCR}}(x_t,c,t),
\end{equation}
with
\begin{equation}
\delta v_{\mathrm{CCR}}(x_t,c,t)
\propto
\frac{s}{t}
\Bigl(
v_{\theta_{\mathrm{ref}}}(\tilde{x}_s,c,s)
-
v_{\theta_{\mathrm{ref}}}(\tilde{x}_s,\varnothing,s)
\Bigr).
\label{eq:app_ccr_v_update}
\end{equation}

\paragraph{Why the second re-noising step matters.}
Both direct distillation and CCR use a reference conditional contrast on top of RL, but they differ in where that contrast is evaluated:
\begin{itemize}
    \item direct distillation uses the original re-noised training state $(x_t,t)$;
    \item CCR uses a second re-noised state $(\tilde{x}_s,s)$ constructed after following the current model prediction.
\end{itemize}
This distinction matters because RL gradually changes the effective denoising geometry of the model. A contrast defined at the original training state remains tied to where the model started that update step. By contrast, CCR first accepts the model's current prediction and then builds the structural correction in the local region induced by that prediction. In this sense, CCR provides a more prediction-conditioned structural prior, rather than repeatedly injecting a reference contrast tied only to the original training location.

\paragraph{A local first-order view.}
Let
\begin{equation}
D(x,s)
=
v_{\theta_{\mathrm{ref}}}(x,c,s)
-
v_{\theta_{\mathrm{ref}}}(x,\varnothing,s)
\end{equation}
denote the frozen reference contrast field. The direct variant uses $D(x_t,t)$, whereas CCR uses $D(\tilde{x}_s,s)$, where
\begin{equation}
\tilde{x}_s
=
(1-s)\bigl(x_t-t\,v_\theta(x_t,c,t)\bigr)+s\epsilon'.
\end{equation}
Since $\tilde{x}_s$ depends on the current prediction, a local first-order expansion implies that $D(\tilde{x}_s,s)$ varies with $v_\theta(x_t,c,t)$ through the induced state $\tilde{x}_s$. In this sense, CCR acts as a prediction-conditioned local structural prior, whereas direct distillation uses a contrast fixed at the original re-noised training state.

\paragraph{Interpretation.}
The goal of CCR is not to force the learned correction to reproduce the pretrained CFG correction at the original training state. Rather, its role is to make it easier for RL to retain an appropriately positive component along the pretrained CFG direction while still following reward-driven optimization. This is why CCR is designed as a local structural prior around the model's current prediction, rather than as a direct target defined only at the original re-noised training state.

\section{KL-Coefficient Sweep}
\label{app:kl_sweep}

This appendix provides additional results for the KL-coefficient sweep in NFT training.
In the main text, we compare the default KL setting $\lambda_{\mathrm{KL}}=10^{-4}$ with a representative larger-KL setting $\lambda_{\mathrm{KL}}=10^{-2}$.
Here we additionally include the intermediate setting $\lambda_{\mathrm{KL}}=10^{-3}$ to verify whether increasing KL regularization yields a consistent improvement in the reward--quality trade-off.

Table~\ref{tab:kl_sweep} shows the results on Z-Image across four reward settings.
Overall, the sweep does not reveal a consistent monotonic benefit from stronger KL regularization.
Increasing the KL coefficient can improve some prompt-alignment metrics or visual-clarity metrics in certain settings, but these gains are uneven and often come with reduced reward or weaker performance on other quality dimensions.
For example, under GenEval, larger KL improves DPG and Laplacian relative to the default setting, but the intermediate and larger settings behave similarly in reward while exhibiting different quality trade-offs.
Under PickScore and HPSv2.1, stronger KL does not improve the target reward and only yields inconsistent quality gains.
Under the multi-reward setting, the default KL coefficient already achieves the best reward, and stronger KL does not provide a clear overall advantage.

These results justify the choice in the main text to report only the default KL setting and a representative stronger-KL setting. More importantly, they show that simply increasing KL strength does not reliably preserve the quality-related structure lost during RL, reinforcing our broader argument that controlling policy drift in magnitude alone is insufficient to preserve the relevant geometric structure.

\begin{table*}[t]
\centering
\small
\setlength{\tabcolsep}{5pt}
\begin{tabular*}{\textwidth}{@{\extracolsep{\fill}} l c c c c c c c c c @{}}
\toprule
\multirow{2}{*}{\textbf{Setting}} 
& \multicolumn{3}{c}{\textbf{Reward $\uparrow$}} 
& \multicolumn{3}{c}{\textbf{Prompt Alignment (DPG) $\uparrow$}} 
& \multicolumn{3}{c}{\textbf{Visual Clarity (Lap.) $\uparrow$}} \\
\cmidrule(lr){2-4}
\cmidrule(lr){5-7}
\cmidrule(lr){8-10}
& $10^{-4}$ & $10^{-3}$ & $10^{-2}$
& $10^{-4}$ & $10^{-3}$ & $10^{-2}$
& $10^{-4}$ & $10^{-3}$ & $10^{-2}$ \\
\midrule
GenEval    & 0.938 & 0.953 & 0.952
           & 79.8  & 82.7  & 83.3
           & 208   & 108   & 331 \\
PickScore  & 0.842 & 0.842 & 0.830
           & 82.8  & 82.0  & 82.2
           & 229   & 240   & 443 \\
HPSv2.1    & 0.377 & 0.373 & 0.350
           & 80.4  & 82.1  & 82.8
           & 928   & 943   & 875 \\
Multi-rew. & 1.456 & 1.456 & 1.407
           & 85.8  & 85.7  & 85.7
           & 755   & 692   & 954 \\
\bottomrule
\end{tabular*}
\caption{KL-coefficient sweep for NFT on Z-Image across four reward settings. We compare $\lambda_{\mathrm{KL}}\in\{10^{-4},10^{-3},10^{-2}\}$. Increasing KL regularization does not yield a consistent improvement in the reward--quality trade-off: stronger KL can improve some prompt-alignment or sharpness metrics in certain settings, but the trend is not stable across rewards or metrics.}
\label{tab:kl_sweep}
\end{table*}

\section{Computation of the Average Projection Coefficient}
\label{app:projection_computation}

This appendix describes how we compute the average projection coefficient used in the geometric analysis. The goal is to measure how much of the learned RL correction remains aligned with the pretrained CFG correction on average across denoising steps. The full procedure is summarized in Algorithm~\ref{alg:avg_projection}.

For each prompt, we first generate a clean latent $x_0$ using the pretrained model. We then sample a shared Gaussian noise $\epsilon$ and construct noisy states
\begin{equation}
x_t=(1-t)x_0+t\epsilon
\end{equation}
for timesteps $t\in\mathcal T$. At each timestep, we compute:
\begin{itemize}
    \item the pretrained conditional and unconditional predictions to form $\Delta_{\mathrm{cfg}}$;
    \item the RL-updated conditional prediction to form $\Delta_{\mathrm{learn}}$ relative to the pretrained conditional prediction.
\end{itemize}
We then compute the stepwise projection coefficient and average it across timesteps.

\begin{algorithm}[t]
\caption{Computation of the average projection coefficient}
\label{alg:avg_projection}
\small
\begin{algorithmic}[1]
\REQUIRE Pretrained model $v_{\theta_{\mathrm{pre}}}$, RL-updated model $v_{\theta_{\mathrm{rl}}}$, prompt set $\mathcal{C}$, denoising timestep set $\mathcal{T}$
\FOR{each prompt $c \in \mathcal{C}$}
    \STATE Generate a clean latent $x_0$ using the pretrained model
    \STATE Sample Gaussian noise $\epsilon \sim \mathcal{N}(0,I)$
    \FOR{each $t \in \mathcal{T}$}
        \STATE Construct $x_t = (1-t)x_0 + t\epsilon$
        \STATE Compute $v_{\theta_{\mathrm{pre}}}(x_t,c,t)$ and $v_{\theta_{\mathrm{pre}}}(x_t,\varnothing,t)$
        \STATE Compute $v_{\theta_{\mathrm{rl}}}(x_t,c,t)$
        \STATE Form
        \[
        \Delta_{\mathrm{cfg}}(x_t,c,t)
        =
        v_{\theta_{\mathrm{pre}}}(x_t,c,t)
        -
        v_{\theta_{\mathrm{pre}}}(x_t,\varnothing,t)
        \]
        \STATE Form
        \[
        \Delta_{\mathrm{learn}}(x_t,c,t)
        =
        v_{\theta_{\mathrm{rl}}}(x_t,c,t)
        -
        v_{\theta_{\mathrm{pre}}}(x_t,c,t)
        \]
        \STATE Compute
        \[
        \alpha(x_t,c,t)
        =
        \frac{\left\langle \Delta_{\mathrm{learn}}(x_t,c,t),\Delta_{\mathrm{cfg}}(x_t,c,t)\right\rangle}
        {\|\Delta_{\mathrm{cfg}}(x_t,c,t)\|_2^2}
        \]
    \ENDFOR
    \STATE Compute
    \[
    \bar{\alpha}(c)
    =
    \frac{1}{|\mathcal{T}|}\sum_{t\in\mathcal{T}}\alpha(x_t,c,t)
    \]
\ENDFOR
\STATE Report the mean of $\bar{\alpha}(c)$ over evaluation prompts
\end{algorithmic}
\end{algorithm}

\section{Guided Baseline with CFG-Style Conditional Masking}
\label{app:guided_training_baseline}

This appendix describes the guided baseline used in the main text. The purpose of this baseline is to test whether simply enabling external guidance during training and inference can mitigate the quality degradation observed after RL, without using the local structural prior introduced by CCR.

\paragraph{Training-time conditional masking.}
Starting from standard NFT-style RL training, we apply classifier-free dropout to the text condition. For each training sample, let $e(c)$ denote the text embedding of prompt $c$. We replace it by
\begin{equation}
\tilde{e}(c)
=
m\cdot e(c),
\qquad
m\sim\operatorname{Bernoulli}(1-p_{\mathrm{drop}}),
\end{equation}
where $m=1$ keeps the conditional embedding and $m=0$ replaces it by the unconditional branch. This masking is applied independently per sample.

The rest of the RL objective remains unchanged. The same masked conditioning is used in the old-policy branch, the current trainable branch, and the frozen reference branch. Thus, this baseline injects unconditional training signals in a CFG-style way, but does not introduce the local prediction-conditioned structural correction used by CCR.

\paragraph{Inference-time guidance.}
At inference time, this baseline uses standard external CFG by combining conditional and unconditional predictions with a nonzero guidance scale. Therefore, unlike CCR, which preserves CFG-related structure through training-time regularization while still using a single conditional forward pass at test time, this baseline continues to rely on external guidance during sampling and thus requires both a conditional and an unconditional forward pass at each denoising step.

\end{document}